\documentclass[runningheads]{llncs}
\usepackage{graphicx}
\usepackage{comment}
\usepackage{color}
\usepackage{multirow}
\usepackage{makecell}
\usepackage{cite}
\usepackage{bm}
\usepackage{amsmath,amssymb}
\newcommand{\etal}{\textit{et al}. }
\newcommand{\argmax}{\mathop{\rm arg~max}\limits}

\begin{document}
\title{Deep Selection: A Fully Supervised Camera Selection Network for Surgery Recordings}
\titlerunning{Deep Selection}

\author{Ryo Hachiuma\inst{1}\orcidID{0000-0001-8274-3710} \and
Tomohiro Shimizu\inst{1} \and
Hideo Saito\inst{1}\orcidID{0000-0002-2421-9862} \and 
Hiroki Kajita\inst{2} \and
Yoshifumi Takatsume\inst{2}}
\authorrunning{R. Hachiuma et al.}

\institute{Keio University, Yokohama, Kanagawa, Japan \\
\{ryo-hachiuma, tomy1201, hs\}@keio.jp \and Keio University School of Medicine, Shinjuku-ku, Tokyo, Japan \\
\{jmrbx767, tsume\}@keio.jp}

\maketitle

\begin{abstract}
Recording surgery in operating rooms is an essential task for education and evaluation of medical treatment. However, recording the desired targets, such as the surgery field, surgical tools, or doctor's hands, is difficult because the targets are heavily occluded during surgery. We use a recording system in which multiple cameras are embedded in the surgical lamp, and we assume that at least one camera is recording the target without occlusion at any given time. As the embedded cameras obtain multiple video sequences, we address the task of selecting the camera with the best view of the surgery. Unlike the conventional method, which selects the camera based on the area size of the surgery field, we propose a deep neural network that predicts the camera selection probability from multiple video sequences by learning the supervision of the expert annotation. We created a dataset in which six different types of plastic surgery are recorded, and we provided the annotation of camera switching. Our experiments show that our approach successfully switched between cameras and outperformed three baseline methods.
\end{abstract}

\section{Introduction}
Recording plastic surgeries in operating rooms with cameras has been indispensable for a variety of purposes, such as education, sharing surgery technologies and techniques, performing case studies of diseases, and evaluating medical treatment \cite{Matsumoto2013Digital, Sadri2013Video}. The targets that best depict the surgery, such as the surgical field, doctor's hands, or surgical tools, should be recorded for these purposes. The recording target differs for each surgery type and the purpose of recording.

However, it is difficult to continuously record these targets without any occlusion. If the camera is attached to the operating room environment, the targets may be occluded by the doctors, nurses, or surgical machines. If the camera is attached to the doctor's head to record video from the first-person view, the camera's field of view does not always capture the target, and the video is often affected by motion blur because of the doctor's head movements. Therefore, a first-person viewpoint camera is not suitable for recording surgery. Moreover, the doctors are disturbed by the camera which is attached to their head during the surgery which requires careful and sensitive movement.

Shimizu \etal \cite{Shimizu2020Surgery} proposed a novel surgical lamp system with multiple embedded cameras to record surgeries. A generic surgical lamp has multiple light bulbs that illuminate the surgical field from multiple directions to reduce the shadows caused by the operating doctors. Hence, Shimizu \etal expected that at least one of the multiple light bulbs would almost always illuminate the surgical field. In the same way, we can expect that at least one of the multiple cameras embedded in the surgical lamp system will always record the target without occlusion.

As the cameras obtain multiple videos of a single surgery, Shimizu \etal also proposed a method to automatically select the image with the best view of the surgical field at each moment to generate one video. By assuming that the quality of the view is defined by the visibility (area size) of the surgical field, they used an image segmentation \cite{li2013pixel} trained for each surgery using manually annotated images. Then, they applied Dijkstra's algorithm to generate a smooth video sequence in which the selected camera does not change frequently over a period of time.

Although their method only focuses on the visibility of the surgical field, the visibility of the field is not always treated as the cue of the best view for recording surgeries. The recording target differs for each surgery type and the purpose of the recording. In some cases, the pose or the motion of the doctor's hands and surgical tools may be the most important, so the best view should not always be determined by the larger surgical field. A method based on image segmentation of predefined target objects is therefore not suitable to select the best view from the multiple videos of the surgical scene.

In this paper, we aim to establish a method to select the best view based on annotated labels by humans, rather than simply using the area size of the surgery field. We address the task in a fully supervised manner, in which the camera selection label is directly given to the prediction model during training, and the predictor learns to map the video sequences to the corresponding selected camera labels. No prior work has been proposed that addresses the task of selecting the best camera view of surgery recordings in a fully supervised manner.

The naive approach to solve this task is to formulate it as a multi-class classification problem, in which the model outputs the selected camera index. However, if the number of cameras in the recording system changes from the training time to the test time, the model cannot be applied at the test time as the number of categories (camera index) will differ. We, therefore, predict the probability of selecting each camera against each frame in the video sequences instead of predicting the selected camera index. When humans create a single video that depicts the surgery the best from multiple video sequences, they consider not only the temporal information of each video sequence but also the intra-camera context. We, therefore, present a deep neural network that aggregates information from multiple video sequences not only sequentially but also spatially (intra-camera direction) to predict the probability of selecting each camera from each frame. 

As there is no dataset available to the public containing surgery recordings via multiple cameras, we record our dataset using the system proposed by Shimizu \etal \cite{Shimizu2020Surgery}. The actual plastic surgeries are recorded at our university's school of medicine. We record six different types of surgery with five cameras attached to the surgical lamp. We validate our proposed model with this dataset, and we quantitatively evaluate our approach with three baseline methods to verify the effectiveness of our approach. The experiments show that our approach can create a video with the best views from multiple video sequences. 

In summary, our contributions are as follows: (1) To the best of our knowledge, this is the first study to address the task of selecting the cameras with the best views from multiple video sequences for the purpose of recording surgery. (2) We propose an end-to-end deep learning network for the selection of cameras with the best views, which aggregates the visual context features sequentially and spatially. (3) We create a dataset of a variety of plastic surgeries recorded with multiple cameras, we provide camera-switching labels via a human expert, and we conduct extensive experimentation from qualitative and quantitative perspectives to show the effectiveness of our method. Please also refer to our accompanying video.

\section{Related Work}
Multiple cameras are used in many situations, such as office environments, sports stadiums, and downtown areas. Although large areas can be recorded without occlusion using multiple cameras, only the necessary information must be selected from the huge number of video sequences recorded by multiple cameras. Self-controlled camera technology, such as automatic viewpoint switching, multiple video summarization, is therefore regarded as an important issue \cite{chen2014autonomous}.

Liu \etal \cite{liu2001automating} presented a rule-based camera switching method inspired by the heuristic knowledge of a professional video editor. Their method successfully switches between the viewpoints of three cameras shooting the speaker and the audience. Doubek \etal \cite{doubek2004cinematographic} recorded moving objects with multiple cameras embedded in an office environment. Cameras were selected based on their score, and a resistance coefficient was introduced so that the cameras were switched only when the score changed significantly. In these methods, camera switching is conducted based on the heuristic knowledge of the professional.

Recently, Chen \etal \cite{chen2018camera} presented a camera selection method in which the rich deep features are extracted from the neural network and online regression to broadcast soccer games. As they stated in the discussion, their method is sensitive to experimental parameters, such as the range of camera duration and the weight of the regularization term. Moreover, their method requires pre-processing and post-processing of the image and camera selection indices. On the other hand, our end-to-end deep neural network directly learns to map the multiple video sequences to the camera selection indices.

The work most relevant to ours is that presented by Shimizu \etal \cite{Shimizu2020Surgery}. They presented a surgery recording system with multiple cameras embedded in the surgery lamp, with the assumption that at least one of the cameras will always record the surgery field. They also proposed a camera selection method that can select one frame among multiple frames obtained by the cameras. They calculated the area size of the surgery field of each frame, and they applied Dijkstra's algorithm to obtain smooth camera selection indices based on the segmentation score. The aim of the current study differs from theirs. As it is not always best to record the entire surgery field, it is difficult to determine the camera selection criterion beforehand. We, therefore, provide the ground truth of camera selection directly to the predictor, and the predictor learns to map the multiple video sequences to the camera selection process in a fully supervised manner.

\section{Approach}
\begin{figure*}[tb]
\begin{center}
\includegraphics[width=0.95\hsize]{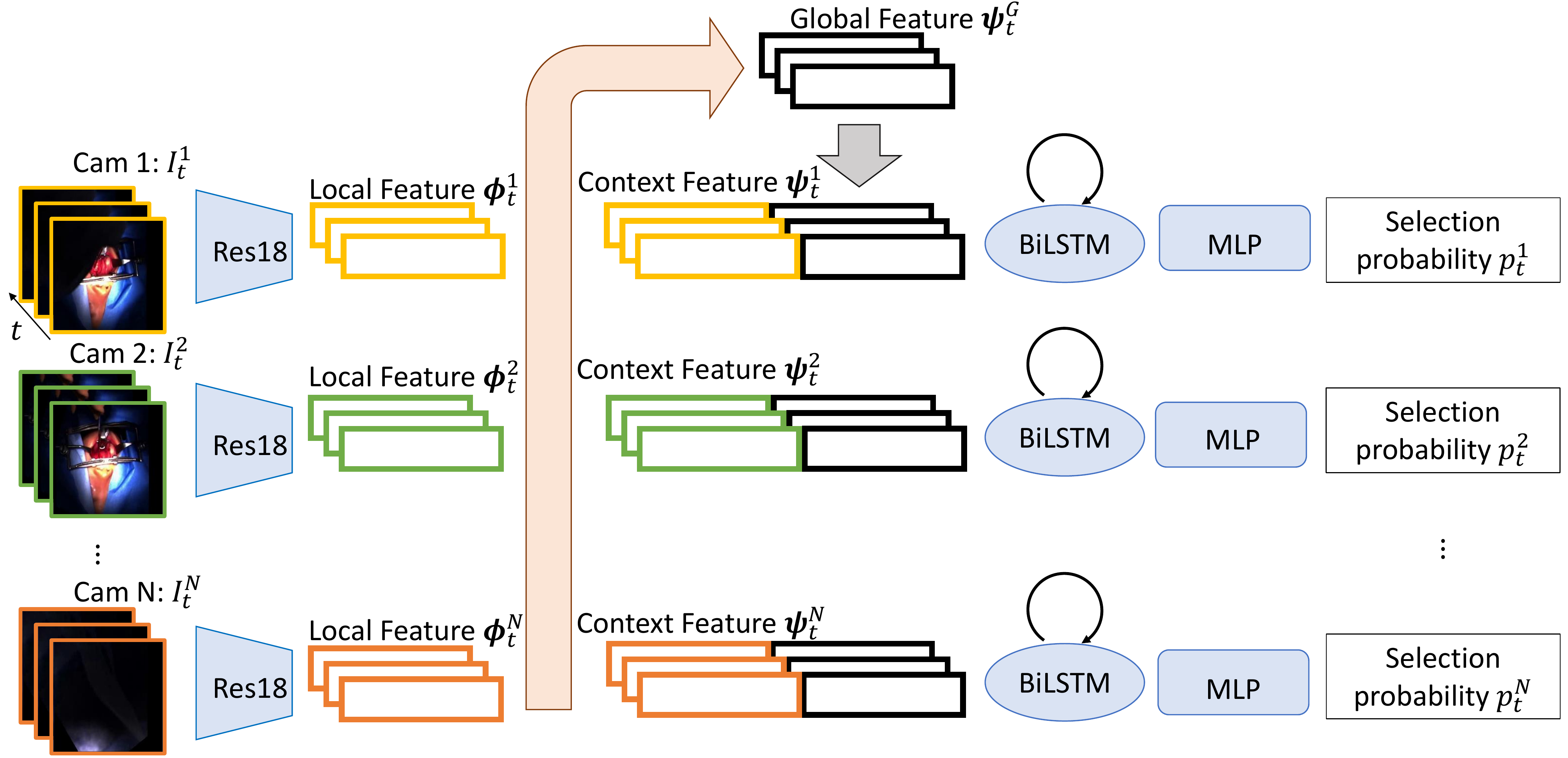}
\end{center}
\caption{The network architecture of the proposed method}
\label{fig:overview}
\end{figure*}
\subsection{Problem Formulation}
Our task is to estimate a sequence of camera labels ${\bf y} = \langle y_1, y_2, \ldots, y_T \rangle$, where $y_t \in [1, N]$, and $T$ denotes the number of frames in the sequence, from image sequences ${\bm  I} = \langle {\bm  I}_1, {\bm  I}_2, \dots, {\bm  I}_T \rangle$, where ${\bm  I}_t = \langle I^1_t, I^2_t, \ldots, I^N_t \rangle$, from $N$ cameras. The naive approach to solve this task is to treat it as an $N$-class classification problem similar to the image classification task: that is to say, to classify a set of images ${\bm  I}_t$ into $N$ classes for each point in time. On the other hand, our network predicts the camera selection probability ${\bm  p}_t^n$ for each camera and solve the task as a binary classification problem. The predicted camera labels $y_t$ can be obtained by calculating the maximum selection probabilities ${\bm  p}_t^n$.

\subsection{Network Architecture}
Our network architecture is represented in Figure \ref{fig:overview}. The network is composed of four components; visual feature extraction, spatial feature aggregation, sequential feature aggregation, and selection probability module. First, we extract the visual feature $\phi_t^n \in {\mathbb R}^{128}$ from each image $I_t^n$. We employ ResNet-18 \cite{He2016ResNet} as the visual feature extractor.

Next, we aggregate the features from multiple cameras at each time step to consider intra-camera (spatial) context, $\psi_t^G = {\mathbb A}(\phi_t^1, \ldots \phi_t^N)$, where ${\mathbb A}$ is a spatial feature aggregation module that combines the independent camera local feature $\phi_t^n$ into an aggregate feature $\psi_t^G$. The aggregated feature should be invariant to the input permutation as the number of cameras or the index of cameras can be changed at the test time. To handle unordered feature aggregation, we use the idea of neural network on 3D point cloud processing. As the order of points should not be mattered for the extraction of the point cloud's global feature, the point cloud's global feature should be invariant to the order of points. Inspired by PointNet \cite{Qi2016Pointnet}, we employ max pooling as ${\mathbb A}$ to extract the global feature $\psi_t^G$. The global feature is concatenated to each local feature $\phi_t^n$ to obtain the context feature $\psi_t^n$.

Then, we aggregate the context features over time, ${\hat \psi}_1^n, \ldots {\hat \psi}_t^n = {\mathbb B}(\psi_1^n, \ldots \psi_t^n)$, where ${\mathbb B}$ is a sequential feature aggregation module that computes the sequential feature ${\hat \psi}_t^n$. In our experiments, we employ a BiLSTM recurrent neural network \cite{Graves2005BiLSTM} with one hidden layer for ${\mathbb B}$ which aggregate not only the past sequence but also future sequence against the target feature.

The output feature ${\hat \psi}_1^n, \ldots, {\hat \psi}_t^n$ is then fed to a multilayer perceptron (MLP) with two hidden layers and leaky rectified linear units (LeakyReLU) activation function \cite{Bing2015Empirical} to predict the selection probability $p_1^n, \ldots, p_T^n$. The sigmoid activation function is applied to the output layer. Final predicted camera labels can be obtained by calculating the maximum selection probabilities at each time step: $y_t = \argmax_{1 \leq n \leq N} p_t^n$ where $p_t^n$ denotes a camera's selection probability at the time step $t$ of $n$-th camera. 

\subsection{Loss function}
Because we formulate the task as binary classification problem, the optimal weight of the network can be obtained using the binary cross entropy between the prediction $p_t^n$ and the ground-truth ${\hat p}_t^n$. However, a class imbalance exists between two categories: selected and not selected. For example, the ratio of selected labels and not-selected labels is $1:3$ in the case of four cameras. This class imbalance
during training overwhelms the cross entropy loss. Easily classified negatives (not selected) comprise the majority of the loss and dominate the gradient. We therefore employ the focal loss \cite{Lin2017Focal}, inspired by the two-stage object detection method. The focal loss $L$ we employ is as follows:
\begin{equation}
L({\bm \xi}) = -\frac{1}{NT}\sum_{n=1}^N \sum_{t=1}^T(1-q_{t,n})^\gamma \log q_{t,n},
\end{equation}
where $q_{t,n}$ is $p_t^n$ when the $n$-th camera at the time step $t$ is selected as the best camera view (${\hat p}_t^n=1$), and $q_{t,n}$ is $1-p_t^n$ when the $n$-th camera at the time step $t$ is not selected as the best camera view (${\hat p}_t^n=0$). ${\bm \xi}$ is the parameters of the proposed model. $\gamma$ is {\it focusing} parameter $\gamma \geq 0$.

\section{Experiments}
\subsection{Dataset}
As there is no dataset available that contains surgery recordings with multiple cameras, we use the system proposed by Shimizu \etal \cite{Shimizu2020Surgery} to create our dataset. The surgeries are recorded at Keio University School of Medicine. Video recording of the patients is approved by the Keio University School of Medicine Ethics Committee, and written informed consent is obtained from all patients or parents. We record six different types of surgery with five cameras attached to the surgical lamp. Each surgery is $30$ minutes long, and the video is recorded at $30$ frames per second (FPS). We subsampled every five frames for each sequence. The ground-truth annotations are created by a single expert.

\subsection{Network training}
We employ Adam optimizer \cite{kingma2015Adam} with a learning rate of $1.0 \times 10^{-4}$. When training the model, we randomly sample a data fragment of $T=40$ frames. The model converged after $50$ epochs, which takes about six hours on a GeForce Quadro GV100. We apply dropout with the $0.5$ probability during training to reduce the overfitting at the first layer of MLP. We set $\gamma=2.0$ of focal loss \cite{Lin2017Focal}. The weights of ResNet-18 is initialized with the pretrained ImageNet \cite{Deng2009Image}. We train the model with batch size $2$.

\subsection{Comparison with the other method and baselines}
No prior work has been proposed that addresses the task of switching between cameras using deep learning for the creation of surgical videos. We therefore set three baseline methods to validate our approach. We compare the following baselines:
\begin{itemize}
    \item \textbf{Ours w/o spa., seq.}: Our camera selection method without spatial feature aggregation and sequential feature aggregation. This method estimates the camera probability using ResNet-18 and MLP.
    \item \textbf{Ours w/o spa.}:  Our camera selection network without the spatial feature aggregation module. In this model, the context of other cameras are not considered to predict each camera's selection probability.
    \item \textbf{Ours w/o seq.}: Our camera selection network without the sequential feature aggregation module. In this model, the sequential context is not considered to predict each camera's selection probability.
\end{itemize}

Moreover, we also compare the results with Shimizu \etal \cite{Shimizu2020Surgery} as the reference. The focus of this paper is to predict the camera label which is adapted for each expert. We do not aim to generate the model generalized for different experts, but the model that represents each expert’s subjective selection. In contrast, Shimizu \etal \cite{Shimizu2020Surgery} aims to select the camera based on the generic criterion obtained by segmenting the surgical field. This means that we aim the different goals from Shimizu \etal \cite{Shimizu2020Surgery}. However, comparing the accuracy of the method \cite{Shimizu2020Surgery} evaluates how much the expert see the surgical field during the annotation. We follow the experimental setup for training the segmentation model and switching parameter as same as the original paper \cite{Shimizu2020Surgery}.

\section{Results}
\begin{table}[tb]
\caption{Quantitative results for camera selection for sequence-out and surgery-out settings. The dice score (F value) is employed as the evaluation metric so a higher value is better. S1 to S6 denote the indices of surgeries. The accuracy of Shimizu \etal \cite{Shimizu2020Surgery} is shown as the reference.}
\label{tab:quant}
\centering
\scalebox{1.15}{
\begin{tabular}{l|cccccc||c}
\hline \Xhline{2\arrayrulewidth}
 & \multicolumn{7}{c}{Sequence-Out} \\ \cline{2-8}
Method   & S1  & S2 & S3 & S4  & S5 & S6 & Average\\ \hline
Shimizu \etal \cite{Shimizu2020Surgery} & {\bf 0.58} & {\bf 0.59} & 0.57 & {\bf 0.45} & 0.34 & {\bf 0.56} & 0.52 \\ \hline
Ours w/o spa., seq.  & 0.45       & 0.39       & 0.98       & 0.28       & 0.36       & 0.36       & 0.47 \\
Ours w/o spa.        & 0.50 & 0.42       & 0.95       & 0.42       & 0.39       & 0.39     & 0.52 \\
Ours w/o seq.        & 0.45       & 0.43 & 0.98       & 0.33       & 0.32       & 0.38       & 0.48 \\
Ours                 & 0.54       & 0.44       & {\bf 0.98} & 0.36 & {\bf 0.46} & 0.44 & {\bf 0.54} \\
\hline \Xhline{2\arrayrulewidth}
\end{tabular}
}
\\
\scalebox{1.15}{
\begin{tabular}{l|cccccc||c}
\hline \Xhline{2\arrayrulewidth}
 & \multicolumn{7}{c}{Surgery-Out} \\ \cline{2-8}
Method   & S1  & S2 & S3 & S4  & S5 & S6 & Average\\ \hline
Shimizu \etal \cite{Shimizu2020Surgery} &0.46 & 0.44&0.20& 0.30&0.33&{\bf 0.52}& 0.38 \\ \hline
Ours w/o spa., seq.  & 0.41       & 0.45       & 0.82       & 0.50       & 0.37       & 0.35       & 0.48 \\
Ours w/o spa.        & {\bf 0.46} & 0.45       & 0.89       & 0.69       & 0.38       & 0.38     & 0.54 \\
Ours w/o seq.        & 0.40       & {\bf 0.48} & 0.88       & 0.56       & 0.40       & 0.38       & 0.52\\
Ours                 & 0.44       & 0.47       & {\bf 0.90} & {\bf 0.70} & {\bf 0.42} & 0.43 & {\bf 0.56} \\ 
\hline \Xhline{2\arrayrulewidth}
\end{tabular}
}
\end{table}
\subsection{Sequence-Out Evaluation}
In this setting, we sequentially split six surgery videos into training and test data at the ratio of $80\mbox{--}20$. In this setup, we adopted the sequence-out cross-validation protocol to evaluate our method. That is to say, the surgery type is known at the test-time but the sequence is unknown at the test-time. Even though the surgery type is known at the test-time, this setting is challenging as single surgery contains multiple sub-tasks so the scene drastically changes between the training and test sequence. The quantitative results are summarized in Table \ref{tab:quant}. As shown in Table \ref{tab:quant}, the prediction accuracy improves with the use of each module in the proposed network. It can be seen that Shimizu \etal \cite{Shimizu2020Surgery} outperformed the other methods for four sequences. However, the rich segmentation annotations are needed at each surgery for Shimizu \etal \cite{Shimizu2020Surgery}.

\begin{figure*}[tb]
\begin{center}
\includegraphics[width=\hsize]{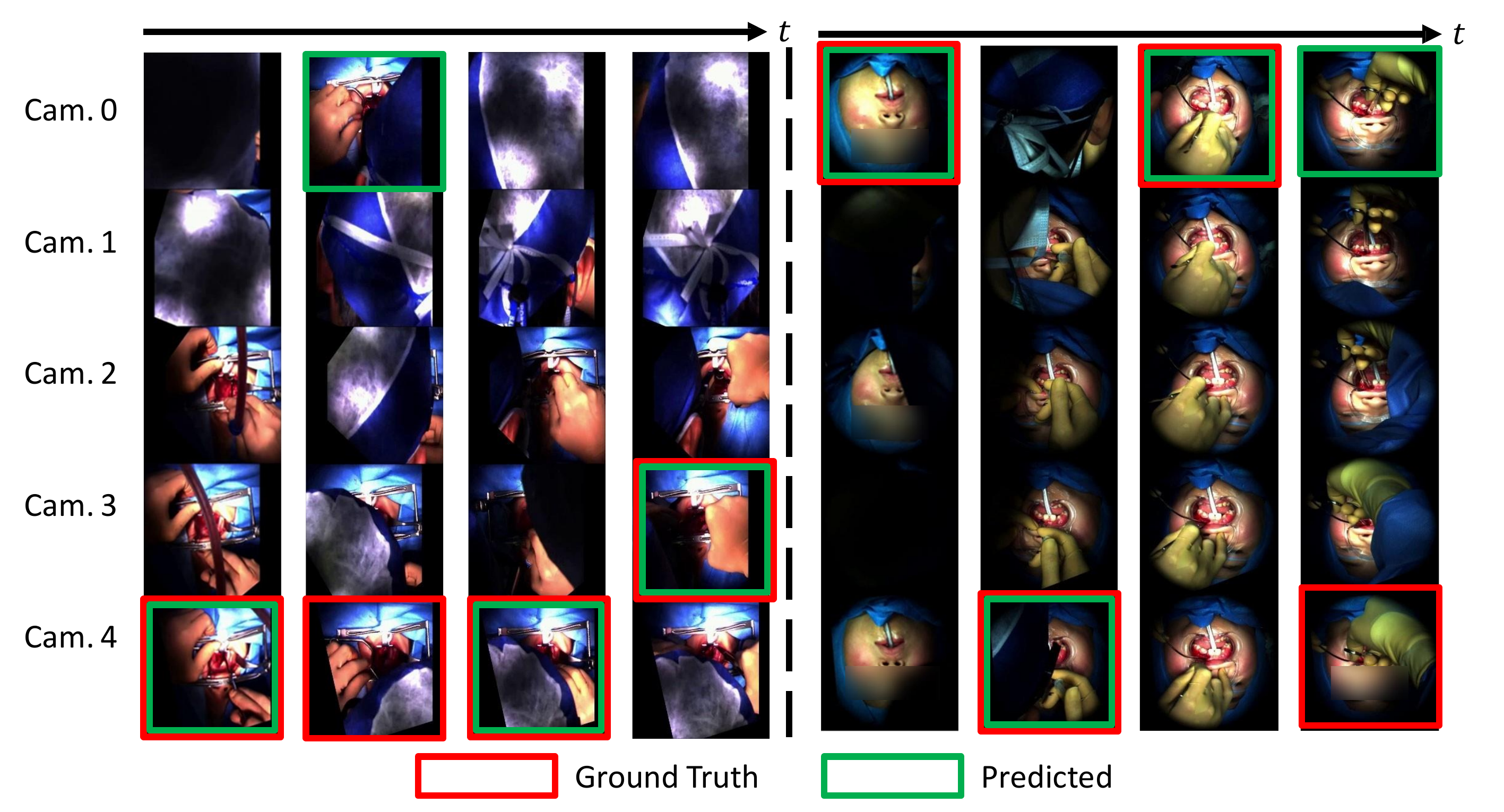}
\end{center}
\caption{Qualitative results for (left) Sequence-Out setting and (right) Surgery-Out setting. The predicted camera frame is highlighted in green, and the ground-truth camera frame is highlighted in red. The column direction indicates the time step, and the row direction indicates the camera indices (Cam.0, \ldots, Cam.4)}
\label{fig:qualt}
\end{figure*}

The qualitative results of camera switching are shown in Figure \ref{fig:qualt} (left). As Figure \ref{fig:qualt} shows, our model successfully selects the frame with the best view from among five cameras, and the predicted camera indices are similar to the camera indices of the ground truth. For example, in the first frame (leftmost column), the surgical field is the most visible at the frame from Camera $4$.

\subsection{Surgery-Out Evaluation}
To further test the robustness of the proposed method, we perform the experiments in the surgery-out setting, in which we train our model on four surgery videos and test it on the two videos. In this setup, we adopted the surgery-out cross-validation protocol to evaluate our method. This is a more challenging setting since the surgery type is completely unknown to the model at the test-time, and a variety of surgery types exist in the dataset. The quantitative results are summarized in Table \ref{tab:quant}. As shown in Table \ref{tab:quant}, our method outperforms other baseline methods. As the unseen surgical field appeared at the test time for Shimizu \etal \cite{Shimizu2020Surgery}, the performance significantly drops for surgery-out settings. 

The qualitative results of camera switching is visualized in Figure \ref{fig:qualt} (right). Even though the type of the surgery is not included in the training dataset, our model successfully selects the best-viewed frame as like the ground-truth's. For the last frame, even though the predicted camera index is not matched to the ground-truth index, the surgery field, surgical tools, and the hand are captured within the frame. 

\section{Conclusion}
We tackled, for the first time, the task of selecting the camera with the best view from multiple video sequences of a surgery. Our model learns to map each video sequence for selection probability while aggregating the features along the intra-camera and temporal directions in a fully supervised manner. Our experiments revealed that our method successfully selects the same camera indices as the ground truth. As the video is divided into the sub-sequences ($40$ frames) to input to the network, the model cannot consider the whole sequence of the long surgery video for the prediction. Therefore, we will investigate the autoregressive model which relies on the prediction of the previous output.

\section*{Acknowledgement}
This research was funded by JST-Mirai Program Grant Number JPMJMI19B2, ROIS NII Open Collaborative Research 2020-20S0404, SCOPE of the Ministry of Internal Affairs and Communications and Saitama Prefecture Leading-edge Industry Design Project, Japan. We would like to thank the reviewers for their valuable comments.

\bibliographystyle{splncs04}

\bibliography{ref}

\begin{thebibliography}{10}
\providecommand{\url}[1]{\texttt{#1}}
\providecommand{\urlprefix}{URL }
\providecommand{\doi}[1]{https://doi.org/#1}

\bibitem{chen2014autonomous}
Chen, J., Carr, P.: Autonomous camera systems: A survey. In: Workshops at the
  Twenty-Eighth AAAI Conference on Artificial Intelligence (2014)

\bibitem{chen2018camera}
Chen, J., Meng, L., Little, J.J.: Camera selection for broadcasting soccer
  games. In: Winter Conference on Applications of Computer Vision (WACV). pp.
  427--435. IEEE (2018)

\bibitem{Deng2009Image}
Deng, J., Dong, W., Socher, R., Li, L.J., Li, K., Fei-Fei, L.: {ImageNet: A
  Large-Scale Hierarchical Image Database}. In: Conference on Computer Vision
  and Pattern Recognition (CVPR). IEEE (2009)

\bibitem{doubek2004cinematographic}
Doubek, P., Geys, I., Svoboda, T., Van~Gool, L.: Cinematographic rules applied
  to a camera network. In: The fifth Workshop on Omnidirectional Vision, Camera
  Networks and Non-Classical Cameras. pp. 17--29. Prague, Czech Republic: Czech
  Technical University (2004)

\bibitem{Graves2005BiLSTM}
Graves, A., Fern\'{a}ndez, S., Schmidhuber, J.: Bidirectional lstm networks for
  improved phoneme classification and recognition. In: Proceedings of the 15th
  International Conference on Artificial Neural Networks: Formal Models and
  Their Applications - Volume Part II. p. 799–804. Springer-Verlag (2005)

\bibitem{He2016ResNet}
{He}, K., {Zhang}, X., {Ren}, S., {Sun}, J.: Deep residual learning for image
  recognition. In: Conference on Computer Vision and Pattern Recognition
  (CVPR). pp. 770--778. IEEE (Jun 2016)

\bibitem{kingma2015Adam}
Kingma, D.P., Ba, J.: Adam: A method for stochastic optimization. In:
  International Conference on Learning Representations (ICLR) (2015)

\bibitem{li2013pixel}
Li, C., Kitani, K.M.: Pixel-level hand detection in ego-centric videos. In:
  Conference on Computer Vision and Pattern Recognition (CVPR). pp. 3570--3577.
  IEEE (Jul 2013)

\bibitem{Lin2017Focal}
{Lin}, T., {Goyal}, P., {Girshick}, R., {He}, K., {Dollár}, P.: Focal loss for
  dense object detection. In: International Conference on Computer Vision
  (ICCV). pp. 2999--3007. IEEE (Oct 2017)

\bibitem{liu2001automating}
Liu, Q., Rui, Y., Gupta, A., Cadiz, J.J.: Automating camera management for
  lecture room environments. In: Proceedings of the SIGCHI conference on Human
  factors in computing systems. pp. 442--449. ACM (2001)

\bibitem{Matsumoto2013Digital}
Matsumoto, S., Sekine, K., Yamazaki, M., Funabiki, T., Orita, T., Shimizu, M.,
  Kitano, M.: Digital video recording in trauma surgery using commercially
  available equipment. Scandinavian journal of trauma, resuscitation and
  emergency medicine  \textbf{21},  27--27 (Apr 2013)

\bibitem{Qi2016Pointnet}
Qi, C.R., Su, H., Mo, K., Guibas, L.J.: Pointnet: Deep learning on point sets
  for 3d classification and segmentation. In: Conference on Computer Vision and
  Pattern Recognition (CVPR). pp. 652--660. IEEE (Jul 2017)

\bibitem{Sadri2013Video}
Sadri, A., Hunt, D., Rhobaye, S., Juma, A.: Video recording of surgery to
  improve training in plastic surgery. Journal of Plastic, Reconstructive \&
  Aesthetic Surgery  \textbf{66}(4),  122 -- 123 (2013)

\bibitem{Shimizu2020Surgery}
Shimizu, T., Oishi, K., Hachiuma, R., Kajita, H., Takatsume, Y., Saito, H.:
  Surgery recording without occlusions by multi-view surgical videos. In:
  International Conference on Computer Vision Theory and Applications (Feb
  2020)

\bibitem{Bing2015Empirical}
Xu, B., Wang, N., Chen, T., Li, M.: Empirical evaluation of rectified
  activations in convolutional network. CoRR  \textbf{abs/1505.00853} (2015),
  \url{http://arxiv.org/abs/1505.00853}

\end{thebibliography}
\end{document}